\documentclass[smallextended]{svjour3}       
\smartqed  
\usepackage{eso-pic}
\newcommand\AtPageUpperMyright[1]{\AtPageUpperLeft{%
		\put(\LenToUnit{0.5\paperwidth},\LenToUnit{-1cm}){%
			\parbox{0.5\textwidth}{\raggedleft\fontsize{9}{11}\selectfont #1}}%
}}%
\newcommand{\conf}[1]{%
	\AddToShipoutPictureBG*{%
		\AtPageUpperMyright{#1}
	}
}
\usepackage{graphicx,amsmath, geometry}
\usepackage{hyperref}
\usepackage{caption}
\usepackage{placeins}
\graphicspath{{./Images/}}
\begin{document}
	\title{ Target State Estimation and Prediction for High Speed Interception }
	\subtitle{}
	\author{Aashay Bhise$^*$ \and Shuvrangshu Jana$^*$ \and Lima Agnel Tony $^*$   \and Debasish Ghose$^*$  }
	\institute{ $^*$ Guidance Control and Decision Systems Laboratory \\
		Department of Aerospace Engineering\\
		Indian institute of Science\\
		Bangalore-12, India\\         
	}
	
	\date{}
	\maketitle
	\conf{MBZIRC Symposium\\ADNEC, Abu Dhabi, 26-27 February 2020}
	\vspace{-2cm}
	\begin{abstract}
		Accurate estimation and prediction of trajectory is essential for interception of any high speed target. In this paper, an extended Kalman filter is used to estimate the current location of target from its visual information and then predict its future position by using the observation sequence. Target motion model is developed considering the approximate known pattern of the target trajectory. In this work, we utilise visual information of the target to carry out the predictions. The proposed algorithm is developed in ROS-Gazebo environment and is verified using hardware implementation. 
	\end{abstract}
	
	\section{Introduction}
	\label{intro}
	Target interception is a challenging problem in the robotic community and its relevance to several applications like defense is increasing day by day. When the target UAV has a speed advantage, its future location estimation becomes difficult. The estimation of target location will aid in precise interception with minimal control effort. The primary motive of performing position estimation and trajectory prediction is to facilitate predictive guidance so that the control effort is optimised. The proposed framework is also robust to intermittent information supply due to the target moving at high speeds. Furthermore, the interception strategy can also be modified if the target is known to follow a fixed trajectory repetitively.    
	
	Several interesting works have been reported in literature about high speed target interception. The improvement of  tracking/interception performance using several methods, is reported in \cite{Ref1}, \cite{Ref2}, \cite{Ref3}. Predictive guidance and learning based guidance are also proposed to improve the interception performance, as seen in \cite{Ref4}, \cite{Ref5}, \cite{Ref6}, \cite{Ref7}. Methods pertaining to the field of soft computing applied in predictive guidance also provide promising results as seen in \cite{Ref8}, \cite{Ref9}, \cite{Ref10}. In existing literature, the target motion model is considered  in general and formulation to include the  known approximate repetitive motion of target using visual information is not reported. 
	In this paper, we present the framework which is designed to estimate and predict the position of a moving target, which follows a repetitive path. Estimation and prediction are essential components in deciding the  point of interception of the target. Essentially, position estimation and trajectory prediction belong to much broader task of approach trajectory generation for moving targets. Two sub-tasks are identified namely, target position estimation and future trajectory prediction of target. While formulating the motion model for target position estimation, the following assumptions regarding the target motion are made. The motion of the target is assumed to be smooth i.e., the change in curvature of the target's trajectory remains bounded and smooth over time. This assumption is the basis of our formulation of the target motion model. The measurement sensor in this case is the vision module. The vision module uses image processing algorithms to compute the estimated position and velocity of the target in inertial plane.  
	In the following section a detailed mathematical formulation of the target state estimation and future state prediction. It is followed by the simulations and on-field results.  
	\section{Mathematical Formulation}
	It is assumed that target is maneuvering on a plane. The position of the target $p_{n}$ and $p_{e}$ are considered as states and it is measured using the vision information. So, the state ($x$) and measurement ($y$) variable  are position of the target and measurement of target information is in camera frame ($p_{e_{\text{image}}}$ and $p_{n_{\text{image}}}$).
	
	\begin{equation}\label{eq:PE_state}
		x = \begin{bmatrix} 
			p_e & p_n \\
		\end{bmatrix} ^T
	\end{equation} 
	
	\begin{equation}\label{eq:PE_measurement}
		y =
		\begin{bmatrix} 
			p_{e_{\text{image}}} & p_{n_{\text{image}}} \\
		\end{bmatrix} ^T
	\end{equation}
	
	\begin{equation}\label{eq:PE_input}
		u = 
		\begin{bmatrix} 
			V_a &
			p_{e_0} &
			p_{n_0} \\
		\end{bmatrix} ^T
	\end{equation}
	An Extended Kalman Filter is used for the target's position estimation. The formulation is done with the consideration that the target's trajectory is lying in the inertial $X-Y$ plane; the $X-Y$ plane being in the E-N direction. The state vector  contains the east and north positions. The input vector ($u$)  contains the estimated speed of the target as given by the vision module and the co-ordinates of the center of the trajectory's curvature. The measurement model \eqref{eq:PE_measurement} contains the position of the target in inertial $X-Y$ plane, as given by the vision module. The co-ordinates of the center of the curvature are calculated by estimating the evolution matrix. Evolution matrix is formulated by writing the future states of the target as a function of current state. The governing relations are as below.
	
	The motion is formulated as in equation \eqref{eq:PE_LS_motion_model} where $r$ is the radius of the instantaneous circle and $\delta$ is the change in $\theta$ between the timesteps. The future states are expressed as a function of previous states as shown in equation \eqref{eq:PE_LS_delta_state}, where $j$ is the index of the observations, and the system of eqns. with the evolution matrix  (\( [ cos\delta \ sin\delta ]^T \)) is as shown in \eqref{eq:PE_LS_matrix_eq}. A sequence of observation is gathered which fills the matrix equation. The Least Squares solution of the observation sequence provides the estimation of evolution matrix at every sample step, so that co-ordinates of the center of curvature, i.e $p_{e_0}$ and $p_{n_0}$ in  \eqref{eq:PE_LS_center}, is available at every time step.    
	
	\begin{equation}\label{eq:PE_LS_motion_model}
		\begin{split}
			p_e(k+1) = p_e(k) - r\delta sin\theta(k)  \\
			p_n(k+1) = p_n(k) + r\delta cos\theta(k)
		\end{split}
	\end{equation}
	
	\begin{equation}\label{eq:PE_LS_delta_state}
		\begin{split}
			\Delta_{p_e}(k,j) = p_e(k-j) - p_e(k-j-1) = -r\delta sin\theta(k-j-1)  \\
			\Delta_{p_n}(k,j) = p_n(k-j) - p_n(k-j-1) = r\delta cos\theta(k-j-1)
		\end{split}
	\end{equation}
	
	\begin{equation}\label{eq:PE_LS_matrix_eq}
		\begin{bmatrix} 
			\vdots  \\
			\Delta_{p_e}(k,j) \\
			\Delta_{p_n}(k,j) \\
			\vdots 
		\end{bmatrix}
		=
		\begin{bmatrix} 
			cos\delta \\
			sin\delta
		\end{bmatrix}
		\begin{bmatrix} 
			\vdots & \vdots \\
			\Delta_{p_e}(k,j-1) & -\Delta_{p_n}(k,j-1) \\
			\Delta_{p_n}(k,j-1) & \Delta_{p_e}(k,j-1)  \\
			\vdots & \vdots
		\end{bmatrix}
	\end{equation}
	
	\begin{equation}\label{eq:PE_LS_center}
		\begin{split}
			p_{e_0} = p_e(k) - \frac{\Delta_{p_n}(k+1)}{\delta} \\
			p_{n_0} = p_n(k) + \frac{\Delta_{p_e}(k+1)}{\delta}
		\end{split}
	\end{equation}
	The EKF formulation and algorithm are very well known. We detail our formulation of prediction model and measurement model to convey important implementation details. Here the goal is to estimate the position in inertial frame (i.e., ENU frame) along the X and Y direction. The 2-dimensional vector \(x\), as shown in equation \eqref{eq:PE_state},  comprises of the positions in inertial frame. A 3-dimensional vector comprising of velocity of target $V_a$ and the instantaneous center co-ordinates in inertial frame ( $p_{n_0}$ and $p_{e_0}$)is fixed as the input vector, as shown in equation \eqref{eq:PE_input}. The process can be described as a non-linear system with,
	\begin{equation}\label{eq:state_eq}
		\dot{x} = F(x,u) + \xi
	\end{equation}
	\begin{equation}\label{eq:PE_state_trans}
		F(x,u) = 
		\begin{bmatrix} 
			- V_a( p_n - p_{n_0} )/\sqrt{( p_n - p_{n_0} ) ^2 + ( p_e - p_{e_0} ) ^2} \\
			V_a( p_e - p_{e_0} )/\sqrt{( p_n - p_{n_0} ) ^2 + ( p_e - p_{e_0} ) ^2} 
		\end{bmatrix} 
	\end{equation}
	where, in equation \eqref{eq:state_eq}, \(F(x,u)\) (refer to equation \eqref{eq:PE_state_trans}) is the non-linear state transition function and \( \xi\) $\sim$ \(\mathcal{N}(0,Q)\) is the process noise where the covariance \(Q\) is generally unknown and can be tuned. The process noise is assumed to be normally distributed. The measurement space contains 2 measurements of \(p_e\) and \(p_n\) in inertial frame, as shown in equation \eqref{eq:PE_measurement}. The measurement model is of the form,
	\begin{equation}\label{eq:meas_eq}
		y = H(x,u) + \eta 
	\end{equation}
	
	\begin{equation}\label{eq:PE_measure_model}
		H(x,u) =
		\begin{bmatrix} 
			p_e\\
			p_n
		\end{bmatrix} 
	\end{equation}
	where, in equation \eqref{eq:meas_eq}, \(H(x,u)\) (refer to equation \eqref{eq:PE_measure_model}) is the non-linear measurement model which maps state and input into measurement space and \( \eta\) $\sim$ \(\mathcal{N}(0,R)\) is the measurement noise where the covariance \(R\) can be estimated by calibrating the sensors. The measurement noise is also assumed to be normally distributed. Prediction step is the first stage of the EKF algorithm where we estimate the state of the system by propagating the previous state and input values to the non-linear process equation \eqref{eq:EKF_prop} in discrete time. The error covariance matrix is projected by the Jacobian of state transition function \eqref{eq:EKF_state_Jac} and perturbed by the process noise covariance, as shown in equation \eqref{eq:EKF_pred_step_cov}.
	
	\begin{equation}\label{eq:EKF_prop}
		\dot{\hat{x}} = F(\hat{x},u) 
	\end{equation}
	
	\begin{equation}\label{eq:EKF_state_Jac}
		A = \frac{\partial F(\hat{x},u)}{\partial x} = \frac{V_a}{(( p_n - p_{n_0} ) ^2 + ( p_e - p_{e_0} ) ^2)^{3/2} }
		\begin{bmatrix} 
			( p_n - p_{n_0} )( p_e - p_{e_0} ) & -( p_e - p_{e_0} )^2 \\
			( p_n - p_{n_0} )^2 & -( p_e - p_{e_0} )( p_n - p_{n_0} ) 
		\end{bmatrix} 
	\end{equation}
	
	\begin{equation}\label{eq:EKF_pred_step_cov}
		\dot{P} = AP + PA^T + Q 
	\end{equation}
	
	The correction step is then carried out after the measurement update where we calculate the Jacobian of the measurement model \eqref{eq:EKF_meas_Jac} and later the Kalman Gain $L$ in \eqref{eq:EKF_kalman_gain}. Then eventually the state estimate \eqref{eq:EKF_state_corr} and the error covariance matrix \eqref{eq:EKF_corr_step_cov} is updated. 
	
	\begin{equation}\label{eq:EKF_meas_Jac}
		C =  \frac{\partial H(\hat{x},u)}{\partial x} =
		\begin{bmatrix} 
			1 & 0\\
			0 & 1
		\end{bmatrix} 
	\end{equation}

	\begin{equation}\label{eq:EKF_kalman_gain}
		L = PC^T(R + CPC^T)^-1 
	\end{equation}
	
	\begin{equation}\label{eq:EKF_corr_step_cov}
		P = (I - LC)P 
	\end{equation}
	
	\begin{equation}\label{eq:EKF_state_corr}
		\dot{\hat{x}} = L(y - H(\hat{x},u)) 
	\end{equation}
	The later part of the task includes predicting the future state of the target based on a sequence of filtered state of target. The position estimation framework provides a filtered position of the target, which is then used for predicting the trajectory of the target. The workflow of trajectory prediction is divided into two, namely, observation phase and prediction phase. During the observation phase, a predefined sequence of observations are gathered \eqref{eq:TP_LS_delta_state}. These observations are the estimated position of the target in inertial frame. Evolution matrix is calculated by obtaining the Least squares solution over the gathered sequence \eqref{eq:TP_LS_matrix_eq}. The evolution matrix is then propagated in the motion model to predict the trajectory in near future. The formulation of the motion model is given in equation \eqref{eq:TP_LS_motion_model}. 
	
	\begin{equation}\label{eq:TP_LS_motion_model}
		\begin{split}
			p_e(k+1) = p_e(k) - (t_{k+1} - t_k)V_a sin\theta(k)  \\
			p_n(k+1) = p_n(k) + (t_{k+1} - t_k)V_a cos\theta(k)
		\end{split}
	\end{equation}
	
	\begin{equation}\label{eq:TP_LS_delta_state}
		\begin{split}
			\Delta_{p_e}(k,j) = p_e(k-j) - p_e(k-j-1) = -(t_{k+1} - t_k)V_a sin\theta(k-j-1)  \\
			\Delta_{p_n}(k,j) = p_n(k-j) - p_n(k-j-1) = (t_{k+1} - t_k)V_a cos\theta(k-j-1)
		\end{split}
	\end{equation}
	
	\begin{equation}\label{eq:TP_LS_matrix_eq}
		\begin{bmatrix} 
			\vdots \\
			\Delta_{p_e}(k,j) \\
			\Delta_{p_n}(k,j) \\
			\vdots
		\end{bmatrix}
		=
		\begin{bmatrix} 
			cos\delta \\
			sin\delta
		\end{bmatrix}
		\begin{bmatrix} 
			\vdots & \vdots \\
			\Delta_{p_e}(k,j-1) & -\Delta_{p_n}(k,j-1)\\
			\Delta_{p_n}(k,j-1) & \Delta_{p_e}(k,j-1) \\
			\vdots & \vdots
		\end{bmatrix}
	\end{equation}
	
	\section{Experimental Results}\label{sec:results}
	\begin{figure}
		\centering
		\includegraphics[scale=0.45]{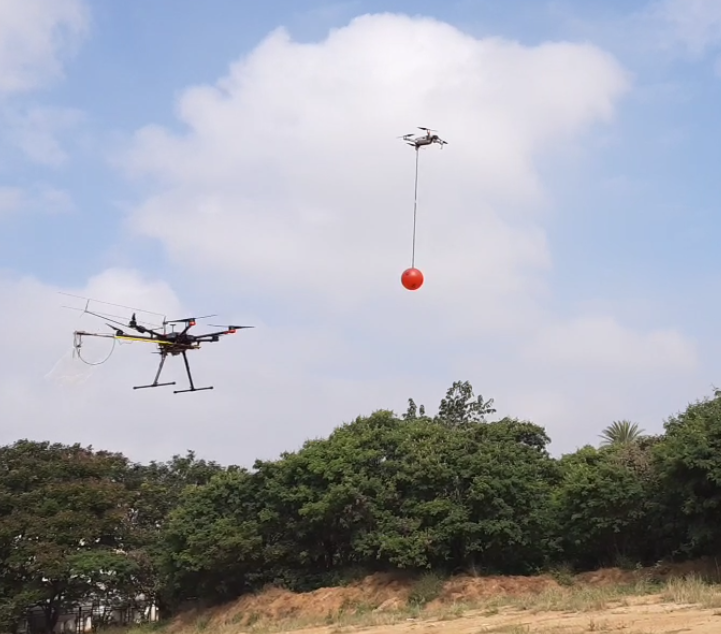}
		\caption{Own UAV and target UAV. Using visual information, own UAV estimates and predicts the location of target}
		\label{fig:UAVs}
	\end{figure}
\begin{figure}
		\centering
		\includegraphics[scale=0.255]{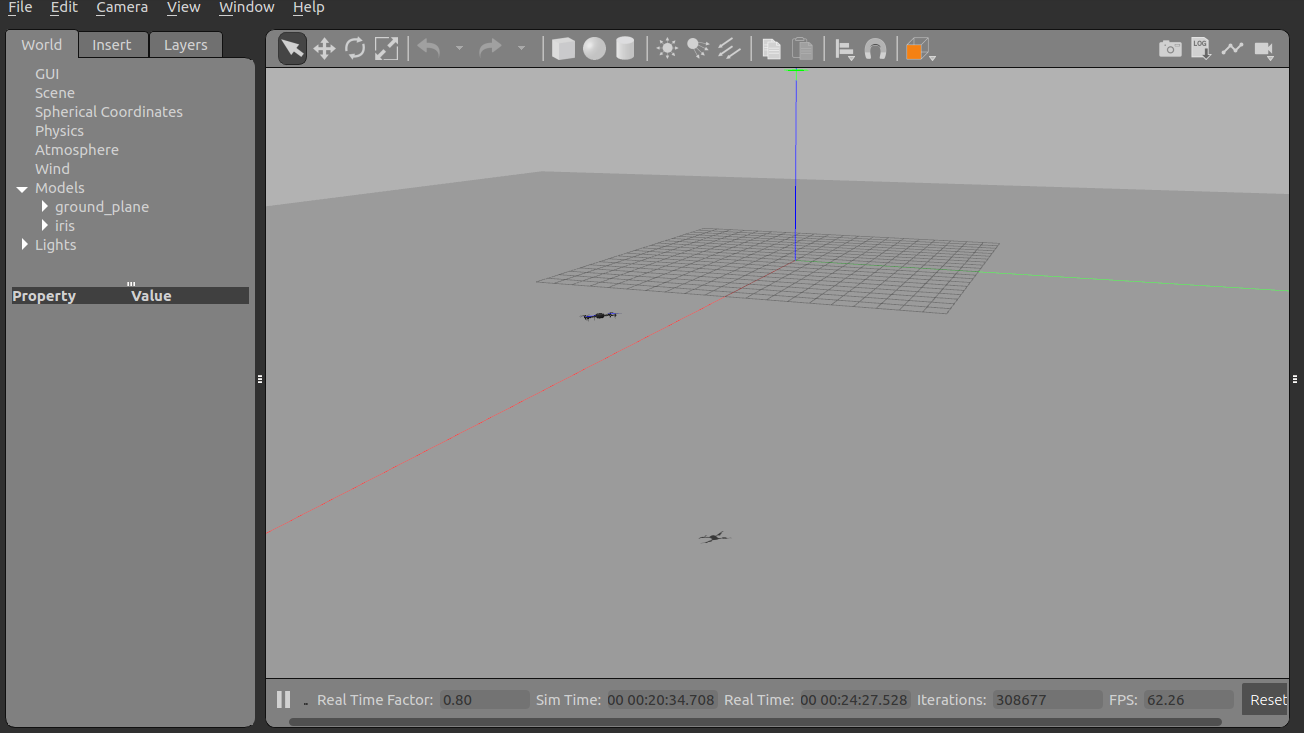}
		\caption{Gazebo environment with IRIS drone}
		\label{fig:gaz_env}
	\end{figure}
\begin{figure}
		\centering
		\includegraphics[scale=0.225]{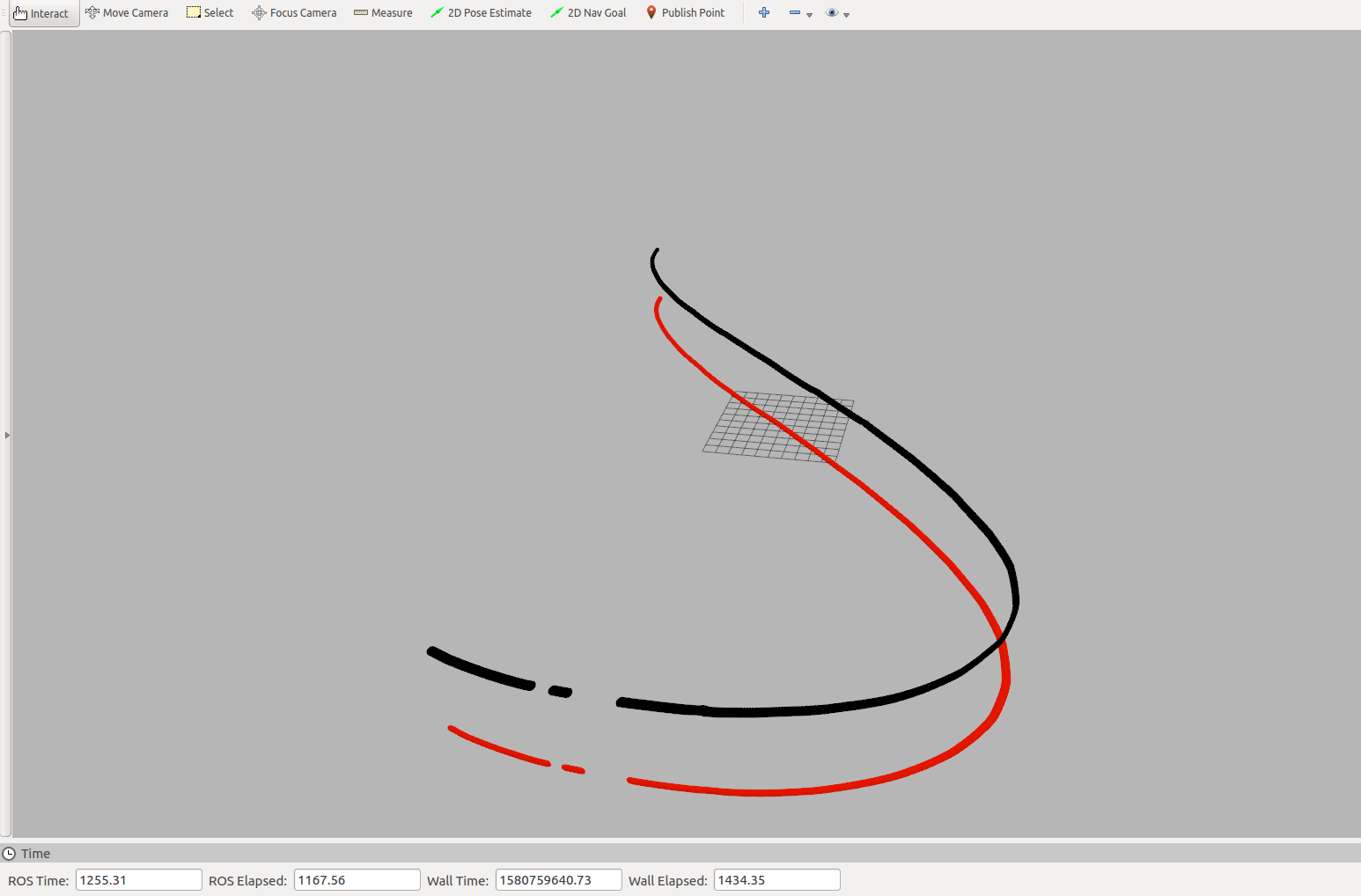}
		\caption{Black trace represents the actual target position and Red trace represents the filtered target position}
		\label{fig:raw_filt_gaz}
	\end{figure}
\begin{figure}
		\centering
		\includegraphics[scale=0.25]{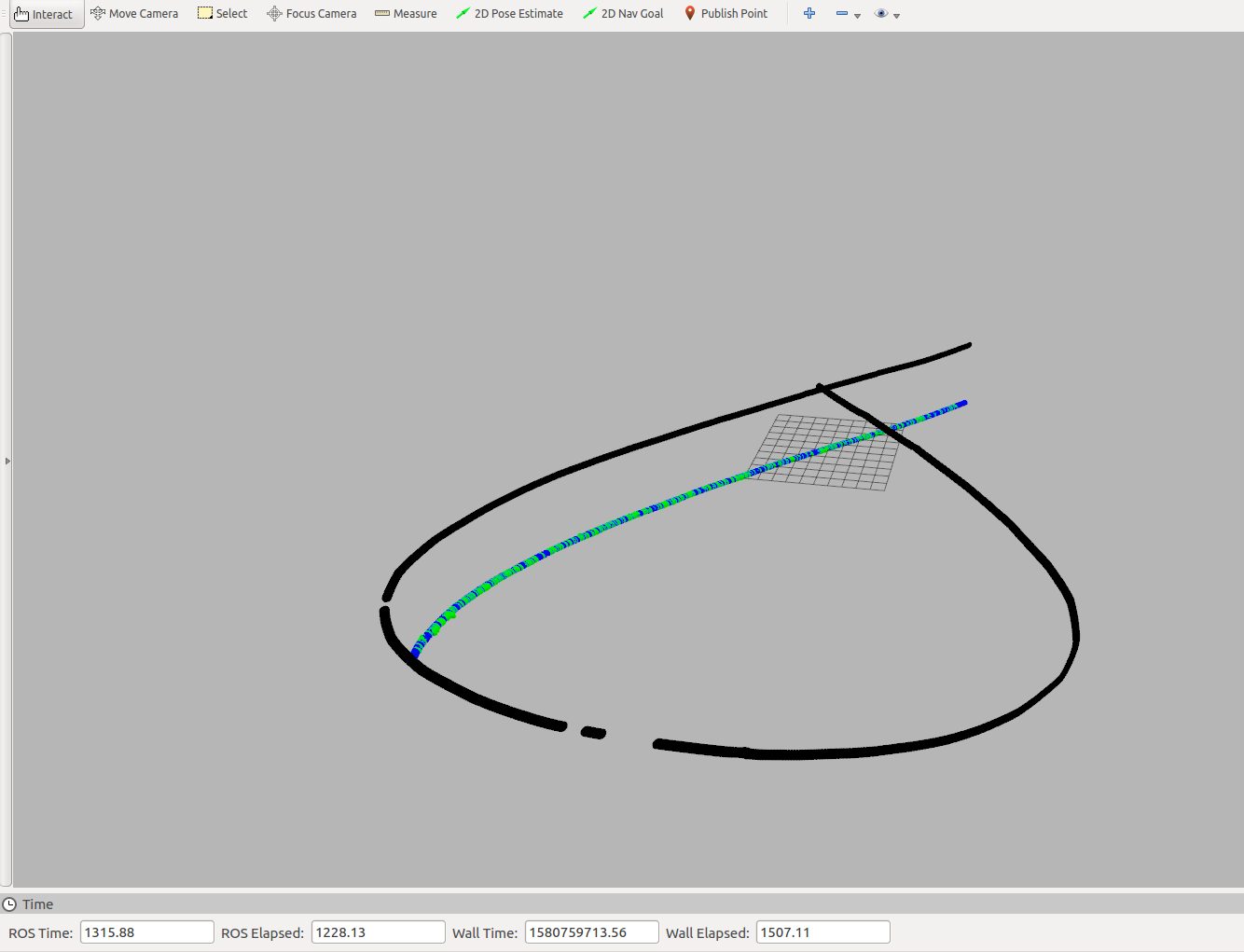}
		\caption{Green trace represents the predicted states and Blue trace represents the states achieved by target}
		\label{fig:gt_pred_gaz}
	\end{figure}
\begin{figure}
		\centering
		\includegraphics[scale=0.225]{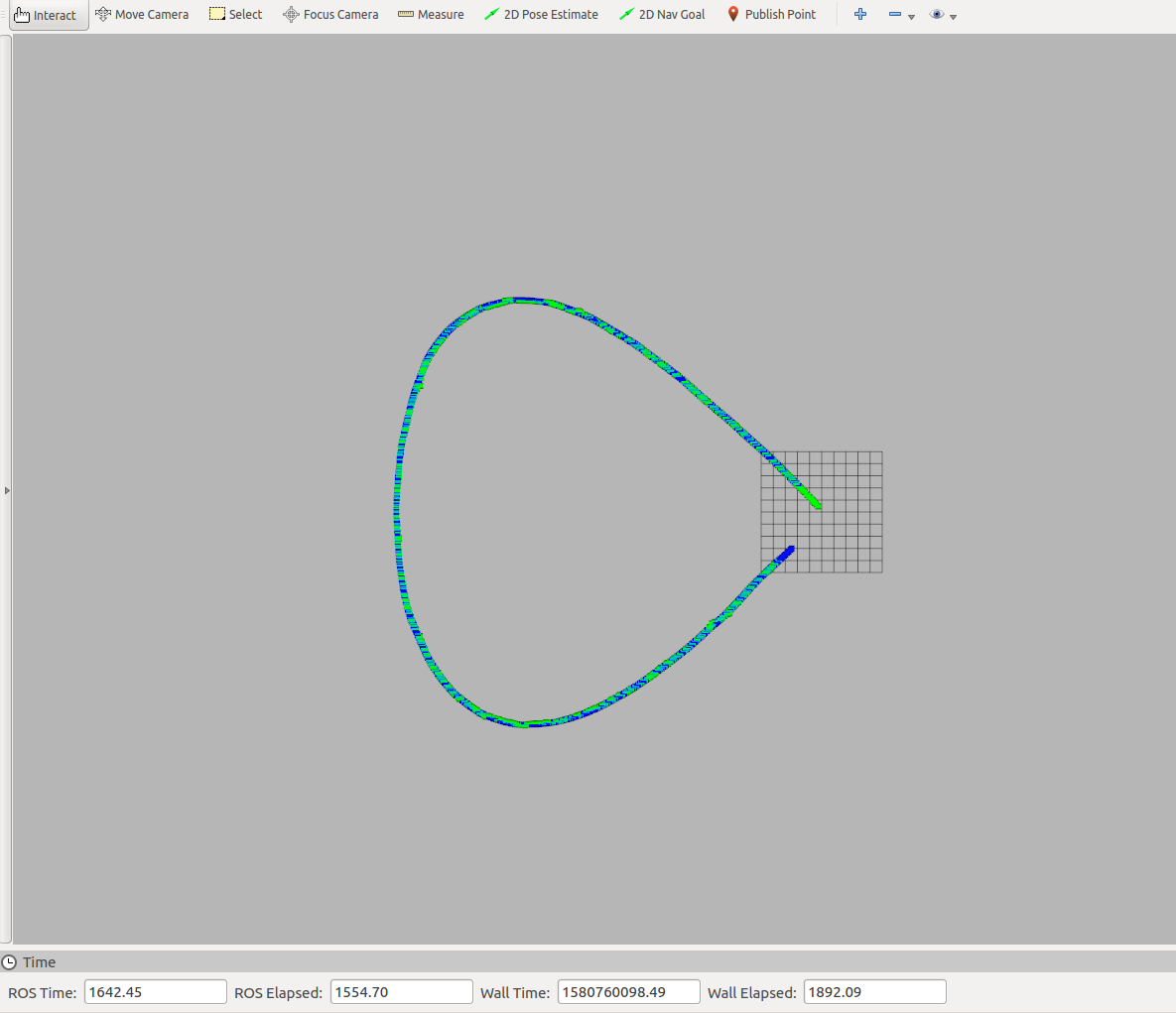}
		\captionsetup{justification=centering}
		\caption{Top down view of the predicted states and achieved states}
		\label{fig:top_down_gaz}
	\end{figure}
\begin{figure}
		\centering
		\includegraphics[scale=0.265]{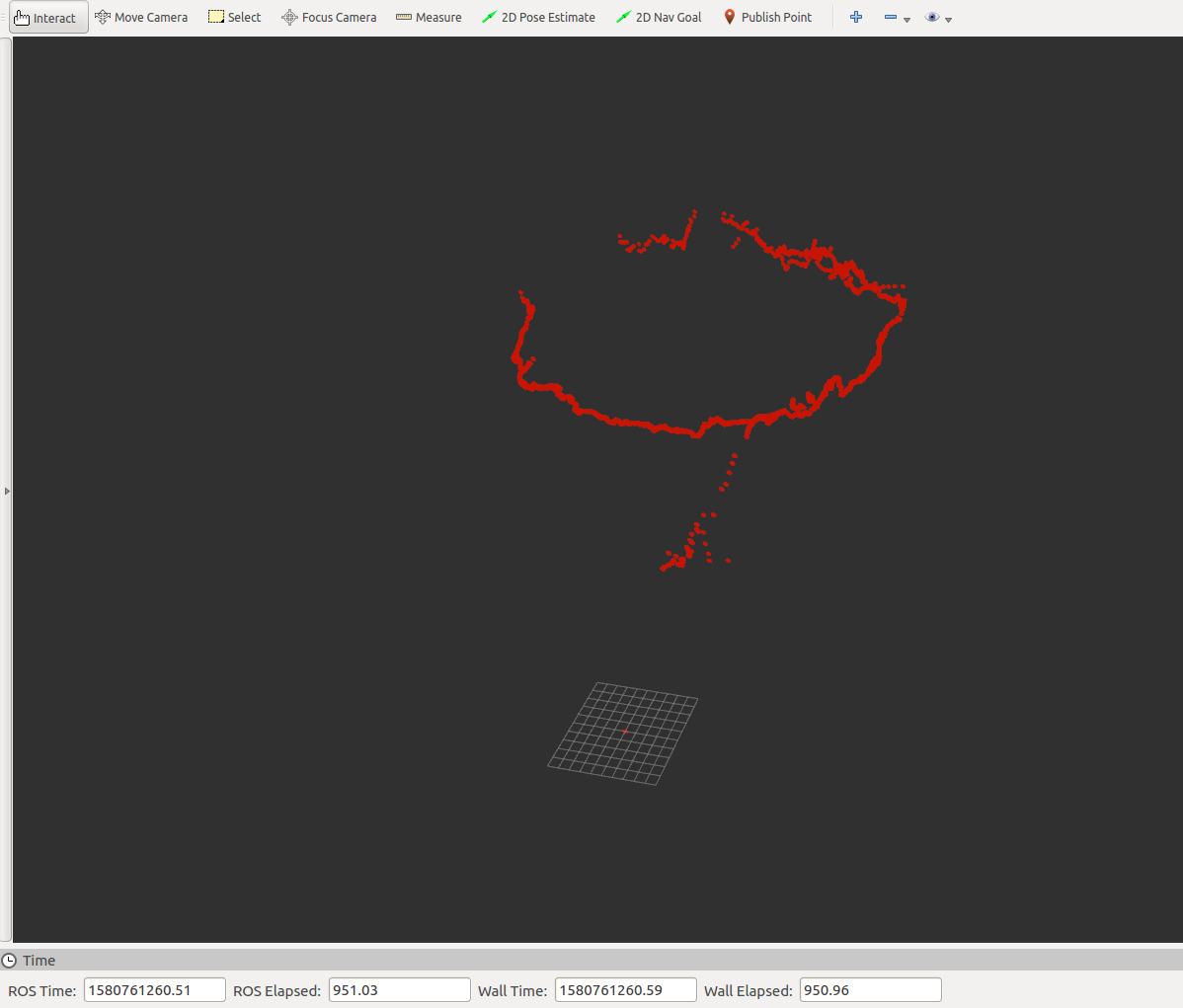}
		\captionsetup{justification=centering}
		\caption{Red trace represents the raw position data as obtained by the tracking the target}
		\label{fig:raw_irl}
	\end{figure}
\begin{figure}
		\centering
		\includegraphics[scale=0.265]{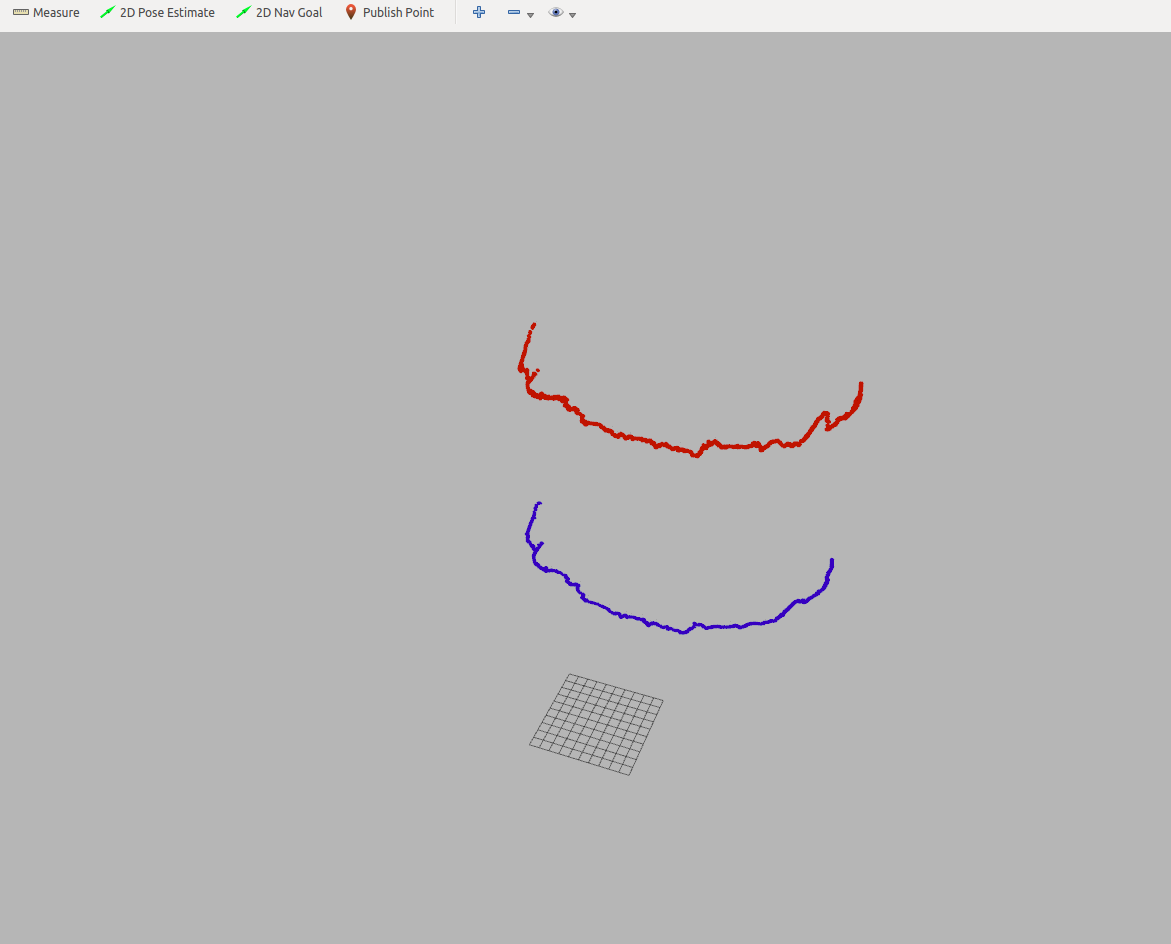}
		\captionsetup{justification=centering}
		\caption{Red trace represents the actual position of target and Blue trace represents the filtered position}
		\label{fig:raw_filt_irl}
	\end{figure}
\begin{figure}
		\centering
		\includegraphics[scale=0.25]{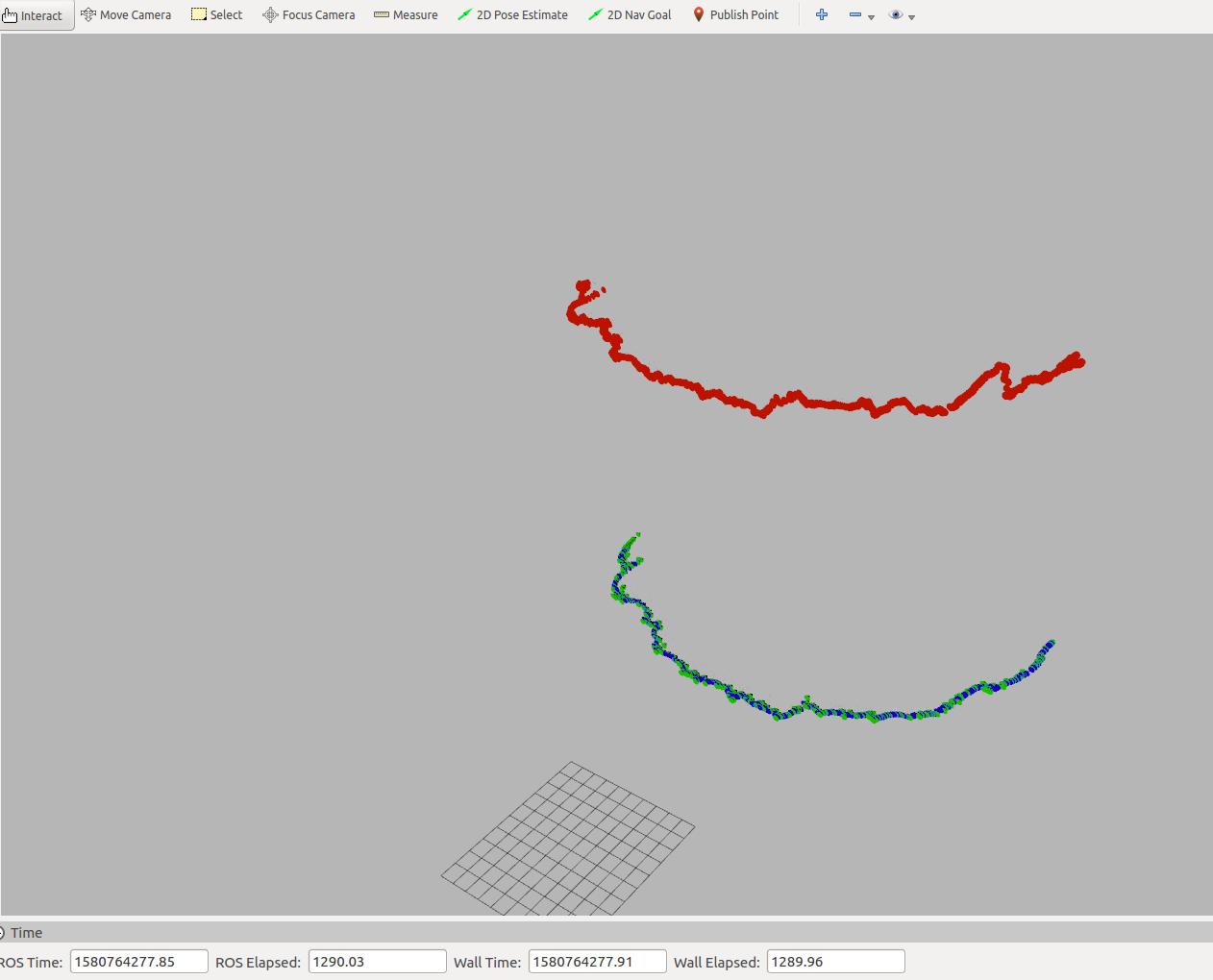}
		\captionsetup{justification=centering}
		\caption{Green trace represents the predicted states and Blue trace represents the states achieved by target}
		\label{fig:gt_pred_irl}
	\end{figure}
\begin{figure}
		\centering
		\includegraphics[scale=0.25]{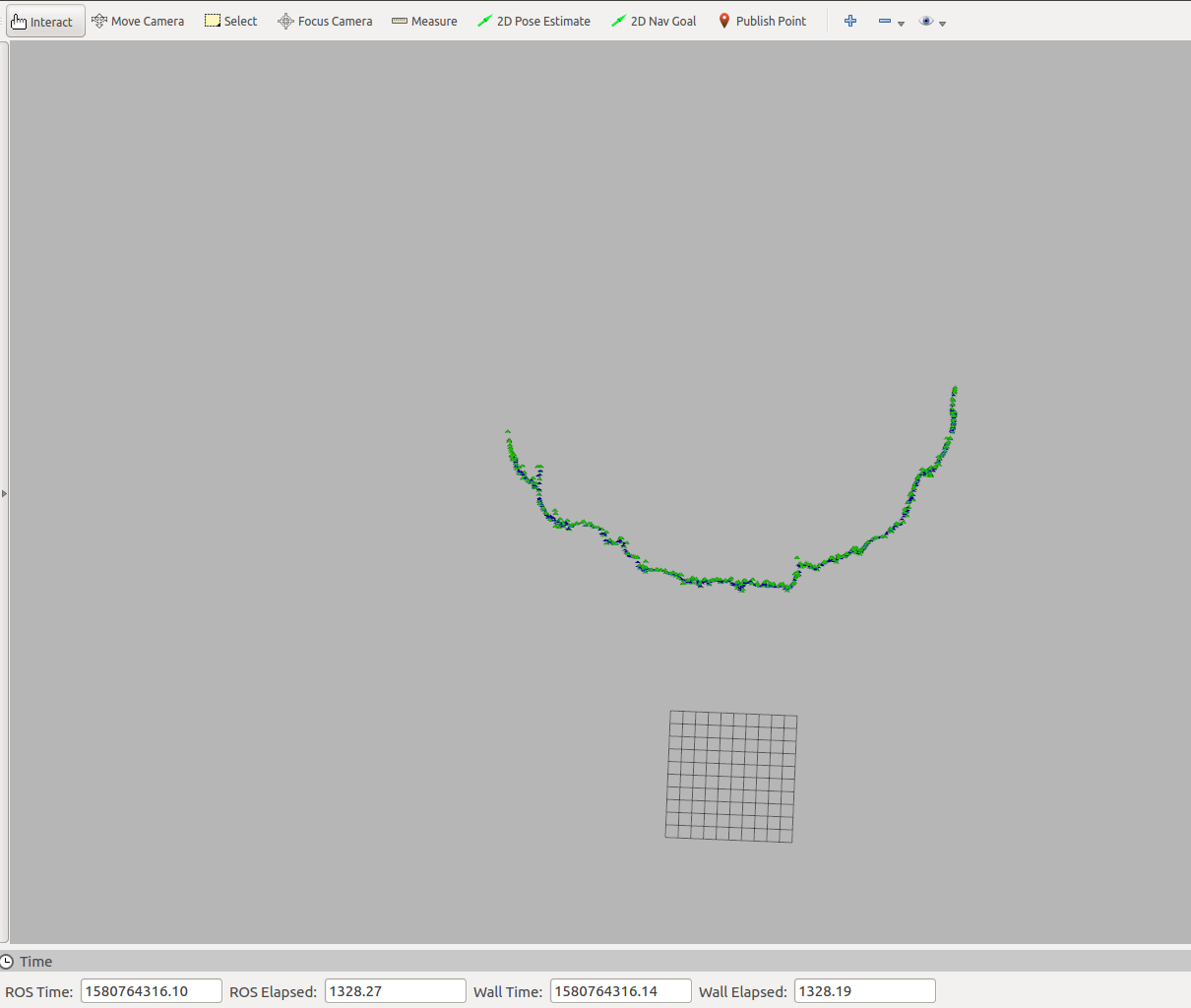}
		\captionsetup{justification=centering}
		\caption{Top down view of the predicted states and achieved states}
		\label{fig:top_down_irl}
	\end{figure}

	A ROS-based pipeline in written in C++ which performs target state filtering and future state estimation. The ROS-based packages are first tested in a gazebo environment where the information about the state of the target is obtained from a simulated drone Fig. \ref{fig:gaz_env}. Since the information about the state of the target is highly accurate and always available, the measurement co-variance is amplified and Gaussian noise is added to observed data, solely to create a realistic scenario. Position estimation of target is done over this noisy data and later, the future state of the target is predicted. A separate process sets the motion of the target in a figure of eight. The states estimated and future states predicted are visualized in RViz. As shown in Fig. \ref{fig:raw_filt_gaz}, the estimated data is visualized against the raw data of the figure of eight curve. Here, since the estimated position is visualized in $X-Y$ plane, we restrict our prediction and estimation in two dimensions. Later, in Fig. \ref{fig:gt_pred_gaz} the ground truth of the targets motion is visualized against the predicted states of the target. By looking at the top down view of the same in Fig. \ref{fig:top_down_gaz}, it can be seen that the ground truth follows the predicted state. 
	A similar on field experiment is done and the states are visualized online in RViz as shown in Fig. \ref{fig:raw_irl}. As shown in Fig. \ref{fig:raw_filt_irl} the estimated data is visualized against the raw data of the target motion. The target motion is rough and non-uniform as it has been subjected to external disturbances, so the estimated states of the target helps in giving a smoother position data. Later, In Fig. \ref{fig:gt_pred_irl} the ground truth of the targets raw motion is visualized against the predicted states of the target. The top down view of the same in Fig. \ref{fig:top_down_irl} is shown.
	
	\FloatBarrier
	\section{Conclusions}
	In this paper, estimation of target location and future state prediction is performed using the  visual information. The proposed method is validated for target motion in circular trajectory. Future work involves the prediction of target location following complex trajectory.
	
	\begin{acknowledgements}
		We would like to acknowledge the Robert Bosch Center for Cyber Physical Systems, Indian Institute of Science, Bangalore, and Khalifa University, Abu Dhabi, for partial financial support. We would also like to thank fellow team members from IISc for their invaluable contributions towards this competition. 
	\end{acknowledgements}

\end{document}